\documentclass[conference]{IEEEtran}
\usepackage{graphicx}
\usepackage{siunitx}
\usepackage{xcolor}
\usepackage{algorithm}
\usepackage{comment}
\usepackage{varwidth}
\usepackage{subfig} 
\usepackage{algpseudocode}
\usepackage{enumitem}
\usepackage{caption}
\usepackage{balance}
\usepackage[backend = bibtex,
			style = numeric,
			sorting = none,
			]{biblatex}
\bibliography{references.bib, caleb-zotero.bib}

\usepackage{url}
\usepackage{multirow}

\newcommand{\todo}[1]{}
\renewcommand{\todo}[1]{{\color{red} TODO: {#1}}}

\newcommand{\eg}[1]{}
\renewcommand{\eg}[1]{(e.g. {#1})}

\newcommand{\ie}[1]{}
\renewcommand{\ie}[1]{(i.e. {#1})}

\newcolumntype{L}[1]{>{\raggedright\let\newline\\\arraybackslash\hspace{0pt}}m{#1}}
\newcolumntype{C}[1]{>{\centering\let\newline\\\arraybackslash\hspace{0pt}}m{#1}}

\begin{document}

\title{A Survey of Methods for Low-Power Deep Learning and Computer Vision}

\author{\IEEEauthorblockN{Abhinav Goel,
Caleb Tung, 
Yung-Hsiang Lu,
and
George K. Thiruvathukal\IEEEauthorrefmark{2}}\IEEEauthorblockA{School of Electrical and Computer Engineering, Purdue University}\IEEEauthorblockA{\IEEEauthorrefmark{2}Department of Computer Science, Loyola University Chicago}
{\{goel39, tung3, yunglu\}@purdue.edu, gkt@cs.luc.edu}
}

\maketitle

\begin{abstract}
Deep neural networks (DNNs) are successful in many computer vision tasks. However, the most accurate DNNs require millions of parameters and operations, making them energy, computation and memory intensive. This impedes the deployment of large DNNs in low-power devices with limited compute resources. Recent research improves DNN models by reducing the memory requirement, energy consumption, and number of operations without significantly decreasing the accuracy. This paper surveys the progress of low-power deep learning and computer vision, specifically in regards to inference, and discusses the methods for compacting and accelerating DNN models. The techniques can be divided into four major categories: (1) parameter quantization and pruning, (2) compressed convolutional filters and matrix factorization, (3) network architecture search, and (4) knowledge distillation. We analyze the accuracy, advantages, disadvantages, and potential solutions to the problems with the techniques in each category. We also discuss new evaluation metrics as a guideline for future research.
\end{abstract}



\section{Introduction}

Deep Neural Networks (DNNs) are widely used in computer vision tasks like
object detection, classification, and segmentation~\cite{ZZ, xliu}.
DNNs are designated as ``Deep" because they 
are made of many layers with a large spread of connections between layers. This gives DNNs a tremendous range of variability that can be fine-tuned for accurate inference through training. 
Unfortunately, DNNs are also computation-heavy and energy-expensive as a result. VGG-16~\cite{vgg}
needs 15 billion operations to perform image classification on a single image~\cite{han2016}. Similarly, YOLOv3 performs
39 billion operations to process one image~\cite{redmon_you_2016}.
These many computations require significant compute resources and lead to a high energy cost~\cite{nguyen_high-throughput_2019}.


This presents a problem for DNNs: how can they be meaningfully deployed on low-power embedded systems and mobile devices? Such machines are often constrained by battery power or obtain energy through low-current USB connections~\cite{mohan}. They also do not usually come with GPUs. Offloading computing to the cloud is a solution~\cite{kumar}, but many DNN applications need to be performed on low-power devices, e.g., computer vision deployed on drones flying in areas without reliable network coverage to offload computation, or in satellites where offloading is too expensive~\cite{anup}. 

Some low-power computer vision techniques remove redundancies from DNNs to reduce the number of operations by 75\% and the inference time by 50\% with negligible loss in accuracy. To deploy DNNs on small embedded computers, more such optimizations are necessary.
Therefore, pursuing low-power improvements in deep learning for efficient inference is worthwhile and is a growing area of research~\cite{lpirc}.

This paper surveys the literature and reports state-of-the-art solutions for low-power computer vision.
We focus specifically on low-power DNN inference, not training, as the goal is to attain high throughput. The paper classifies the low-power inference methods into four categories: 
\begin{enumerate}[noitemsep,nolistsep]
    \item Parameter Quantization and Pruning: Lowers the memory and computation costs by reducing the number of bits used to store the parameters of DNN models. 
    
    \item Compressed Convolutional Filters and Matrix Factorization: Decomposes large DNN layers into smaller layers to decrease the memory requirement and the number of redundant matrix operations.
    
    \item Network Architecture Search: Builds DNNs with different combinations of layers automatically to find a DNN architecture that achieves the desired performance.
    
    \item Knowledge Distillation: Trains a compact DNN that mimics the outputs, features, and activations of a more computation-heavy DNN.
    
\end{enumerate}

TABLE~\ref{tab:comp} summarizes these methods. 
This survey will focus on the above mentioned software-based low-power computer vision techniques,
without considering low-power hardware optimizations \eg{hardware accelerators, spiking DNNs}. 
This paper uses results reported in the existing literature to analyze the advantages, disadvantages, and propose potential improvements to the four methods. We also suggest an additional set of evaluation metrics to guide future research.

\begin{table*}[t!]
\centering
\begin{tabular}{|C{2.7cm}|L{4.7cm}|L{3.7cm}|L{4.5cm}|}
\hline
Technique & \multicolumn{1}{c|}{Description} & \multicolumn{1}{c|}{Advantages} & \multicolumn{1}{c|}{Disadvantages} \\ \hline
Quantization and Pruning & Reduces precision/completely removes the redundant parameters and connections from a DNN. & Negligible accuracy loss with small model size. Highly efficient arithmetic operations. & Difficult to implement on CPUs and GPUs because of matrix sparsity. High training costs. \\ \hline
Filter Compression and Matrix Factorization & Decreases the size of DNN filters and layers to improve efficiency. & High accuracy. Compatible with other optimization techniques. & Compact convolutions can be memory- inefficient. Matrix factorization is computationally expensive. \\ \hline
Network Architecture Search & Automatically finds a DNN architecture that meets performance and accuracy requirements on a target device. & State-of-the-art accuracy with low energy consumption. & Prohibitively high training costs. \\ \hline
Knowledge Distillation & Trains a small DNN with the knowledge of a larger DNN to reduce model size. & Low computation cost with few DNN parameters. & Strict assumptions on DNN structure. Only compatible with softmax outputs. \\ \hline
\end{tabular}
\caption{Comparison of different techniques for performing low-power computer vision.}
\label{tab:comp}
\vspace{-0.3cm}
\end{table*}





\section{Parameter Quantization and Pruning}
Memory accesses contribute significantly to the energy consumption of DNNs~\cite{han2016, flightnn}. To build low-power DNNs, recent research has looked into the tradeoff between accuracy and the number of memory accesses. 


\subsection{Quantization of Deep Neural Networks}
\begin{figure}[b!]
	\vspace{-0.7cm}
	\centering
		\includegraphics[width=0.49\textwidth]{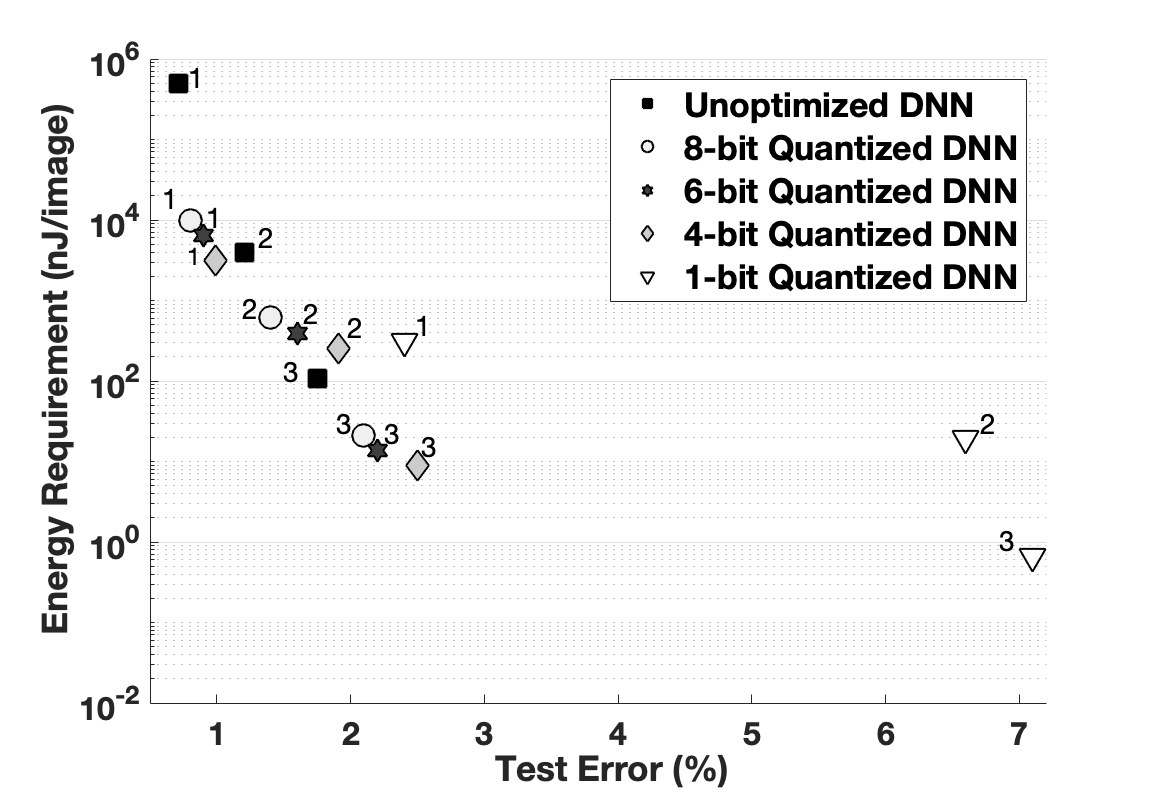}
		\caption{Pareto analysis of DNNs with varying levels of quantization on the MNIST dataset. Lower-left is better because it indicates low energy and smaller error. The annotations $^{1,2,3}$ on the data-points represent three different DNN architectures for each quantization technique used in the experiments~\cite{Goel2018}.}
	\label{fig:par}
\end{figure}

One method to reduce the number of memory accesses is to decrease the size of DNN parameters.
Some methods~\cite{Wang2018,cour2015} show that it is possible to have negligible accuracy loss even when the precision of the DNN parameters is reduced. Courbariaux et al.~\cite{cour2015} experiment with parameters stored in different fixed-point formats to demonstrate that reduced bit-widths are sufficient for training DNNs. Fig.~\ref{fig:par} compares the energy consumption and test error of different DNN architectures with varying levels of quantization. Here, as the parameter bit-width decreases, the energy consumption decreases at the expense of increasing test error. Building on these findings, LightNN~\cite{lightnn}, CompactNet~\cite{Goel2018}, and FLightNN~\cite{flightnn} find the optimal bit-width for different parameters of a DNN, given an accuracy constraint. Moons et al.~\cite{moons2017} also use DNNs with parameters in different integer formats. Binarized neural networks proposed in Courbariaux el al.~\cite{BinNet} and Rastegari et al.~\cite{xnor} train DNNs with binary parameters and activations. In binarized neural networks each parameter is represented with a single bit. Because of the major reduction in precision, these DNNs require many layers to obtain high accuracy. In Fig.~\ref{fig:par}, for a given DNN architecture, the 1-bit quantization (binarized neural networks) consumes the least energy and has the highest error. To improve the accuracy of binarized DNNs, Zhou et al.~\cite{dorefa} quantize the back propagation gradients for better training convergence.


Often, parameter quantization is used along with compression to further reduce the memory requirement of DNNs. Han et al.~\cite{han2016} first quantize the parameters into discrete bins. Huffman coding is then used to compress these bins to reduce the model size by approximately 89\%, with negligible accuracy loss. Similarly, HashedNet~\cite{hash} quantizes the DNN connections into hash buckets such that all connections grouped to the same hash bucket share a single parameter. However, because these techniques have a high training cost, their adoption is limited.


\textbf{Advantages:} 
When the bit-widths of the parameters decrease, the prediction performance of DNNs remains constant~\cite{cour2015, robust}. This is because constraining parameters has a regularization effect in the training process. Moreover, when designing custom hardware for DNNs, quantization allows power-hungry multiply-accumulate operations to be replaced with shift~\cite{Goel2018} or XNOR~\cite{xnor} operations: leading to a reduction in circuit area and energy requirements. \textbf{Disadvantages and Potential Improvements:} DNNs employing quantization techniques need to be retrained multiple times, making the training process very expensive~\cite{dorefa}. The training cost must be reduced to make these techniques easier more practical. Furthermore, different layers in DNNs are sensitive to different features. A constant bit-width for all layers can lead to poor performance~\cite{flightnn}. In order to select a different parameter precision for each connection of the DNN (depending on its importance to the output), the precision value can be represented in a differentiable manner and be included in the training process. Thus, during training, each connection will learn its parameter value and the parameter precision. 

\subsection{Pruning Parameters and Connections}

Removing the unimportant parameters and connections from DNNs can reduce the number of memory accesses.
The Hessian-weighted distortion measure can be used to identify the importance of parameters in a DNN~\cite{choi2017}. Some techniques~\cite{damage,surgeon} use this measure to remove redundant parameters and reduce the DNN size. These measure-based pruning methods only operate on the fully-connected layers. 

\begin{table}[b!]
\vspace{-0.1cm}
\centering
\begin{tabular}{|c|r|r|r|r|r|}
\hline
Method & \multicolumn{1}{c|}{LeNet 5} & \multicolumn{1}{c|}{\begin{tabular}[c]{@{}c@{}}LeNet\\ 300-100\end{tabular}} & \multicolumn{1}{c|}{AlexNet} & \multicolumn{1}{c|}{VGG-16} & \multicolumn{1}{c|}{\begin{tabular}[c]{@{}c@{}}Training\\ Time\end{tabular}} \\ \hline
Unoptimized & 1.000 & 1.000 & 1.000 & 1.000 & \textbf{1.000} \\ \hline
P & 0.080 & 0.080 & 0.090 & 0.075 & 4.000 \\ \hline
P+Q & 0.031 & 0.030 & 0.037 & 0.032 & 6.000 \\ \hline
P+Q+C & \textbf{0.025} & \textbf{0.025} & \textbf{0.028} & \textbf{0.020} & 6.000 \\ \hline
\end{tabular}
\caption{Comparison of model compression rates with different DNNs~\cite{han2016}. Note that training time increases as models are compressed. \textbf{P}: Pruning, \textbf{Q}: Quantization, \textbf{C}: Compression.}
\label{tab:comp1}
\end{table}

To extend pruning to convolutional layers, Anwar et al.~\cite{anwar} use particle filtering to locate the pruning candidates. Polyak et al.~\cite{poly} use sample input data and prune the sparsely activated connections. Han et al.~\cite{Shan} use a new loss function to learn both parameters and connections in DNNs. Yu et al.~\cite{nisp} use an algorithm that propagates importance scores to measure the importance of each parameter with respect to the final output. By performing pruning, quantization, and encoding, Deep Compression~\cite{han2016} reduces the model size by~$95\%$. Path-level pruning is also seen in tree-based hierarchical DNNs~\cite{scalable, FastBalanced}. Although these techniques can identify the unimportant connections, they create unwanted sparsity in DNNs. Sparse matrices require special data structures (unavailable in deep learning libraries), and are difficult to map onto modern GPUs. To overcome this issue, some methods~\cite{Li2016,less} concentrate on building pruned DNNs with sparsity constraints.


\textbf{Advantages:}
As seen in TABLE~\ref{tab:comp1}, pruning can be combined with quantization and encoding for significant performance gains. When the three techniques are used together, the VGG-16 model size decreases to $2\%$ of its original size. Furthermore, pruning reduces the complexity of DNNs, and thus reduces overfitting. \textbf{Disadvantages and Potential Improvements:} The training effort associated with DNN pruning is considerable because DNNs have to be pruned and trained multiple times. TABLE~\ref{tab:comp1} shows that the training time increases by 600\% when using pruning and quantization together. This problem is exacerbated when the DNNs are pruned with sparsity constraints~\cite{wen}. Also, the advantages of pruning are noticed only when using custom hardware or special data structures for sparse matrices~\cite{wen}. Channel-level pruning is a potential improvement to the existing connection-level pruning techniques, because it can be performed without any special data structures and does not create unintended matrix sparsity. By developing techniques to automatically identify unimportant channels, it is possible to perform channel-level pruning without a significant training overhead.


\section{Convolutional Filter Compression and Matrix Factorization}

Convolution operations contribute to the bulk of the computations in DNNs, and the fully connected layers contain around 89\% of the parameters in DNNs like AlexNet~\cite{less}. To reduce the power consumption of DNNs, researchers have focused on reducing the computations in convolution layers, and  the number of parameters in fully connected layers.

\subsection{Convolutional Filter Compression}

\begin{table}[b!]
\vspace{-0.3cm}
\centering
\begin{tabular}{|c|r|r|r|}
\hline
Method & \multicolumn{1}{c|}{\begin{tabular}[c]{@{}c@{}}ImageNet\\ Top-1 Acc\end{tabular}} & \multicolumn{1}{c|}{\begin{tabular}[c]{@{}c@{}}Number of\\ Parameters\end{tabular}} & \multicolumn{1}{c|}{\begin{tabular}[c]{@{}c@{}}Number of\\ Operations\end{tabular}} \\ \hline
AlexNet~\cite{Alex} & 57.20\% & 60.00 M & 727 M \\ \hline
SqueezeNet 1.0~\cite{SQN} & 57.50\% & \textbf{1.24 M} & 837 M \\ \hline
SqueezeNet 1.1~\cite{SQN} & 58.00\% & \textbf{1.24 M} & 360 M \\ \hline
MobileNet v3 Large~\cite{Mob} & \textbf{75.20}\% & 5.40 M & 219 M \\ \hline
MobileNet v3 Small~\cite{Mob}& 67.40\% & 2.50 M & \textbf{56 M} \\ \hline
ShiftNet-A~\cite{shiftt} & 70.10\% & 4.10 M & 1,400 M \\ \hline
Shift Attention Layer~\cite{hacene} & 71.00\% & 3.30 M & 538 M \\ \hline
\end{tabular}
\caption{Comparison of convolutional filter compression techniques (accuracy, number of parameters and operations).}
\label{tab:conv}
\end{table}

Smaller convolution filters have considerably fewer parameters and lower computation costs than larger filters. For example, a 1$\times$1 filter has only 11\% parameters of a 3$\times$3 filter. However, removing all large convolution layers affects the translation invariance property of a DNN and lowers its accuracy~\cite{MBA}. Some studies identify and replace redundant filters with smaller filters for DNN acceleration. SqueezeNet~\cite{SQN} is one such technique that uses three strategies to convert $3\times 3$ convolutions into $1\times 1$ convolutions to reduce the number of parameters. TABLE~\ref{tab:conv} compares the performance of different convolutional filter compression techniques: SqueezeNet has 98\% ($1- \frac{1.24}{60}$) fewer parameters than AlexNet, at the expense of a greater number of operations.
MobileNets~\cite{Mob} and SqueezeNet~1.1 reduce the number of operations along with the number of parameters~\cite{lpirc}. MobileNets use depthwise separable convolutions along with bottleneck layers to decrease the computation, latency, and the number of parameters. MobileNets achieve high accuracy by maintaining a small feature size and only expanding to a larger feature space when performing the depthwise separable convolutions.

\textbf{Advantages:} Bottleneck convolutional filters reduce the memory and latency requirements of DNNs significantly. For most computer vision tasks, these techniques obtain state-of-the-art accuracy. Filter compaction is orthogonal to pruning and quantization techniques. The three techniques can be used together to further reduce energy consumption. \textbf{Disadvantages and Potential Improvements:} It has been shown that the $1 \times 1$ convolutions are computationally expensive in small DNNs, and lead to poor accuracy~\cite{shuffle}. It is also difficult to implement depthwise separable convolutions efficiently because their arithmetic intensity (ratio of number of operations to memory accesses) is too low to efficiently utilize the hardware~\cite{shiftt}. The arithmetic intensity of depthwise separable convolutions can be increased by managing memory more effectively. By optimizing the spatial and temporal locality of the parameters in the cache, the number of memory accesses can be reduced.

\subsection{Matrix Factorization}

Tensor decompositions and matrix factorizations represent the DNN operations in a sum-product form for acceleration~\cite{jader,dent}. Such techniques factorize multi-dimensional tensors (in convolutional and fully-connected layers) into smaller matrices to eliminate redundant computation.
Some factorizations accelerate DNNs up to 4$\times$ because they create dense parameter matrices and avoid the locality problem of non-structured sparse multiplications~\cite{wen}.
To minimize the accuracy loss, matrix factorizations are performed one layer at a time. After factorizing the parameters of one layer, subsequent layers are then factorized based on some reconstruction error. The layer-by-layer optimization approach makes it difficult to scale these techniques to large DNNs because the number of factorization hyper-parameters increases exponentially with DNN depth. To apply these methods in large DNNs, Wen et al.~\cite{wen} enforce compact kernel shapes and depth structures to reduce the number of factorization hyper-parameters.

There are multiple matrix factorization techniques. Kolda et al.~\cite{kolda} show that most factorization techniques can be applied to DNNs for acceleration. However, some techniques do not necessarily provide the optimal tradeoff between accuracy and computation complexity~\cite{kolda}. Canonical Polyadic Decomposition (CPD) and Batch Normalization Decomposition (BMD) are the best performing decompositions in terms of accuracy, while the Tucker-2 Decomposition and the Singular Value Decomposition (SVD) result in poor accuracy~\cite{hayashi, tai}. CPD compresses the DNN more than BMD, and thus accelerates the DNN to a greater extent. The accuracy obtained with BMD is higher than CPD. Moreover, the optimization problem associated with CPD is sometimes unsolvable, making the factorization impossible~\cite{tai}. On the other hand, BMD is a stable factorization and it always exists.

\textbf{Advantages:} Matrix factorizations are methods to reduce the computation costs in DNNs. The same factorizations can be used in both convolutional layers and fully connected layers. The performance gain when using CPD and BMD is significant, with small accuracy loss. \textbf{Disadvantages and Potential Improvements:} Because of the lack of theoretical understanding, it is difficult to say why some decompositions (e.g. CPD and BMD) obtain high accuracy, while other decompositions (e.g. Tucker-2 Decomposition and SVD) do not. Furthermore, the computation costs associated with matrix factorization often offset the performance gains obtained from performing fewer operations. Matrix factorizations are also difficult to implement in large DNNs because the training time increases exponentially with increasing depth. The high training time is mainly attributed to the fact that the search space for finding the correct decomposition hyper-parameters is large. Instead of searching through the entire space, the hyper-parameters can be included in the training process and be learned to accelerate training for large DNNs.
\section{Network Architecture Search}

There are many different DNN architectures and optimization techniques to consider when designing low-power computer vision applications. It is often difficult to manually find the best DNN for a particular task when there are many architectural possibilities. Network Architecture Search (NAS) is a technique that automates DNN architecture design for various tasks. NAS uses a Recurrent Neural Network (RNN) controller and uses reinforced learning to compose candidate DNN architectures. These candidate architectures are trained and then tested with the validation set. The validation accuracy is used as a reward function to then optimize the controller's next candidate architecture. NASNet~\cite{NAS} and AmoebaNet~\cite{amoeba} demonstrate the effectiveness of NAS to obtain state-of-the-art accuracy.
To automatically find efficient DNNs for mobile devices, Tan et al.~\cite{MNas} propose MNasNet. This technique uses a multi-objective reward function in the controller to find a DNN architecture that achieves the desired accuracy and latency (when deployed on a target mobile device) requirements. MNasNet is $2.3\times$ faster than NASNet with $4.8\times$ fewer parameters and $10\times$ fewer operations. Moreover, MNasNet is also more accurate than NASNet. Despite the remarkable results, most NAS algorithms are prohibitively computation-intensive, requiring to train thousands of candidate architectures for a single task. MNasNet requires 50,000 GPU hours to find an efficient DNN architecture on the ImageNet dataset.

To reduce the computation costs associated with NAS, some researchers propose to search for candidate architectures based on proxy tasks and rewards, such as working with a smaller dataset (e.g. CIFAR-10) or approximating the device latency with the number of DNN parameters. FBNet~\cite{FB} is one such technique that uses a proxy task (optimizing over a smaller dataset) to find efficient architectures $420\times$ faster than MNasNet. Cai et al.~\cite{proxyless} show that DNN architectures optimized on proxy tasks are not guaranteed to be optimal on the target task, especially when hardware metrics like latency are approximated with the number of operations or the number of parameters. They also propose Proxyless-NAS to overcome the limitations with proxy-based NAS solutions. Proxyless-NAS uses path-level pruning to reduce the number of candidate architectures and a gradient-based approach for handling objectives like latency to find an efficient architecture in approximately 300 GPU hours. A technique called Single-Path NAS~\cite{single} reduces the architecture search time to 4~hours. This speedup comes at the cost of reduced accuracy. 

\textbf{Advantages:} NAS automatically balances the trade-offs between accuracy, memory, and latency by searching through the space of all possible architectures without any human intervention. NAS achieves state-of-the-art performance in terms of accuracy and energy efficiency on many mobile devices. \textbf{Disadvantages and Potential Improvements:} The computational demand of NAS algorithms makes it difficult to search for architectures that are optimized for large datasets. To find an architecture that meets the performance requirements, each candidate architecture must be trained (to check accuracy) and run on the target device (to check latency/energy) to generate the reward function. The time taken to train and measure the performance of each candidate architecture is significant - leading to high computation costs. To reduce the training time, the candidate DNNs can be trained in parallel with different subsets of the data. The gradients obtained from the different data subsets can be merged to produce one trained DNN. However, such parallel training techniques generally result in low accuracy. Accuracy can be increased by using adaptive learning rates while maintaining high convergence rates.
\section{Knowledge Transfer and Distillation}


Large DNNs are more accurate than small DNNs because the greater numbers of parameters allow large DNNs to learn complex functions~\cite{bengi}. Training small DNNs to learn such functions is challenging with the conventional back propagation algorithm. However, some methods train small DNNs to learn complex functions by making the small DNNs mimic larger pre-trained DNNs. These techniques transfer the ``knowledge'' of a large DNN to a small DNN through a process called Knowledge Transfer (KT). Some early techniques utilizing KT~\cite{ba, bucila} have been widely used to perform DNN compression. Here, the small DNN is trained on data labeled by a large DNN in order to learn complex functions. The key idea behind such techniques is that the data labeled by the large DNN contains a lot of information that is useful for the small DNN. For example, if the large DNN outputs a moderately high probability to multiple categories for a single input image, then it might mean that those categories might share some visual features. By forcing the small DNN to mimic these probabilities, the small DNN learns more than what is available in the training data~\cite{cho}.

Hinton et al.~\cite{hinton} propose another class of techniques called Knowledge Distillation (KD), where the training process is significantly simpler than KT based techniques. In their work, the small DNN is trained using a student-teacher paradigm. The small DNN is the student, and an ensemble of specialized DNNs is the teacher. By training the student to mimic the output of the teacher, the authors show that the small DNN can perform the task of the ensemble with some loss in accuracy. To improve the accuracy of the small DNN, Li et al.~\cite{Li} minimize the Euclidean distance of feature vectors between the teacher and the student. Similarly, FitNet~\cite{fitnet} builds small and thin DNNs by making each layer in the student mimic a feature map in the teacher.
However, the metrics used in Li et al.~\cite{Li} and FitNet~\cite{fitnet} require strict assumptions on the structure of the student and are not sufficient to build energy-efficient student DNNs. 
To solve this problem and improve generalizability, Peng et al.~\cite{Peng} utilize the correlation between the metrics as the optimization problem during training.

\textbf{Advantages:} KT- and KD-based techniques can reduce the computation costs of large pre-trained DNNs significantly. Prior research has shown that the concepts used in KD can be used outside computer vision as well~\cite{cho} \eg{semi-supervised learning, domain adaptation, etc.} \textbf{Disadvantages and Potential Improvements:} KD often places strict assumptions on the structure and the size of the student and the teacher, making it difficult to generalize to all applications. Moreover, the current KD techniques rely heavily on the softmax output and do not work with different output layers. Instead of making the student just mimic the outputs of neurons and layers from the teacher, the student can learn the sequence in which the neurons are activated. This removes the requirement that the student and the teacher have the same DNN structure (thus improving generalizability) and reduces the reliance on softmax output layers.

\section{Discussion}
\subsection{Guidelines for Low-Power Computer Vision}

There is no single technique to build efficient DNNs for computer vision. Most techniques are complementary and can be used together for better energy efficiency. We include some general guidelines to consider for low-power computer vision.

\begin{enumerate}
    \item Quantization and reduced parameter precision decrease the model size and the complexity of arithmetic operations significantly, but unfortunately, it is difficult to manually implement quantization in most machine learning libraries. NVIDIA's TensorRT library provides an interface for such optimizations.
    \item When optimizing large pre-trained DNNs, pruning and model compression are effective options. 
    \item When training a new DNN from scratch, compressed convolutional filters and matrix factorizations should be used to reduce the model size and computation.
    \item NAS finds DNNs that are optimized for individual devices with performance guarantees. DNNs with several branches (e.g. Proxyless-NAS, MNasNet, etc.) frequently require expensive kernel launches and synchronizations on GPUs and CPUs.
    \item Knowledge distillation should be used with small or medium-sized datasets. This is because fewer assumptions on the DNN architectures of the student and the teacher are required, leading to higher accuracy.
\end{enumerate}

\subsection{Evaluation Metrics}

Low-power DNNs for computer vision need to be evaluated on more aspects beyond just accuracy. We list some of the major metrics that should be considered.
\begin{enumerate}
\item The test accuracy should be evaluated on large datasets like ImageNet, CIFAR, COCO, etc. K-folds cross-validation is necessary if the training dataset is small.
\item The number of parameters is generally associated with the memory requirement of the DNN. It is important to compare both these metrics when working with quantization and pruning techniques.
\item The number of operations should be evaluated to find the computation costs. When using low precision DNNs, the cost of each operation reduces. In this case, it is important to also measure the energy consumption.
\item The number of parameters and operations are not always proportional to the energy consumption of the DNN~\cite{lpirc}. To find the energy consumption, DNNs should be deployed on a device connected to a power meter.
\end{enumerate}

\section{Summary And Conclusions}

DNNs are powerful tools for performing many  computer vision tasks.
However, because the best-performing (most accurate) DNNs are designed for situations where computational resources are readily available, they are difficult to deploy on embedded and mobile devices. There has been significant research that focuses on reducing the energy consumption of these DNNs with minimal accuracy loss to make them better suited for low-power devices. This survey paper investigates the research landscape for low-power computer vision and identifies four categories of techniques: Quantization and Pruning, Filter Compression and Matrix Factorization, Network Architecture Search, and Knowledge Distillation. These techniques have their own strengths and weaknesses, with no clear winner. Continued research on improving the state-of-the-art low-power techniques will make computer vision deployable on embedded and mobile devices in the future.


\printbibliography

\end{document}